\documentclass[letterpaper, 10 pt, conference]{ieeeconf}  

\IEEEoverridecommandlockouts                              

\usepackage{graphicx}
\usepackage{xcolor}
\usepackage{bm}
\usepackage{amsmath,amssymb,mathrsfs,bm}
\usepackage{tikz}

\usepackage{cite}
\usepackage{soul}
\usepackage{subfigure}
\usepackage{tabularx}
\usepackage[normalem]{ulem}
\usepackage[bookmarks=true]{hyperref}
\usepackage{xurl}
\usepackage{mathtools}
\usepackage{flushend}

\DeclareMathOperator*{\argmax}{arg\,max}

\usepackage{algorithm}
\usepackage{algpseudocode}
\algdef{S}[FOR]{ForEach}[1]{\algorithmicforeach\ #1\ \algorithmicdo}
\algnewcommand\algorithmicinput{\textbf{Input:}}
\algnewcommand\INPUT{\item[\algorithmicinput]}
\algnewcommand\algorithmicoutput{\textbf{Output:}}
\algnewcommand\OUTPUT{\item[\algorithmicoutput]}
\algrenewcommand\algorithmicindent{1.0em}%

\usepackage{eso-pic}
\newcommand\AtPageUpperMyright[1]{\AtPageUpperLeft{%
 \put(\LenToUnit{0.5\paperwidth},\LenToUnit{-1.5cm}){%
     \parbox{0.5\textwidth}{\raggedright\fontsize{11}{11}\selectfont #1}}}}
\newcommand{\conf}[1]{\AddToShipoutPictureBG*{\AtPageUpperMyright{#1}}}

\title{\LARGE \bf
Context-Generative Default Policy for Bounded Rational Agent 
}

\author{Durgakant Pushp, Junhong Xu, Zheng Chen and Lantao Liu 
\thanks{
    The authors are with the Luddy School of Informatics, Computing, and Engineering  at Indiana University, Bloomington, IN 47408, USA. \newline
    E-mail:        {\tt\small \{dpushp, xu14, zc11, lantao\}@iu.edu}. \newline
    This work was supported by NSF \#2006886 and \#2047169.
    } \vspace{-20pt}
}

\conf{\hspace*{-9cm}\textcolor{gray}{This paper has been accepted for publication at the IEEE/RSJ International Conference on Intelligent Robots and Systems \textcolor{blue}{\href{https://iros2024-abudhabi.org/}{(IROS)}}, Abu Dhabi, UAE, 2024. }}

\begin{document}

\maketitle
\thispagestyle{empty}
\pagestyle{empty}

\begin{abstract} 
Bounded rational agents often make decisions by evaluating a finite selection of choices, typically derived from a reference point termed the `default policy,' based on previous experience. 
However, the inherent rigidity of the static default policy presents significant challenges for agents when operating in unknown environment, that are not included in agent's prior knowledge.
In this work, we introduce a context-generative default policy 
that leverages the region observed by the robot to predict unobserved part of the environment, thereby enabling the robot to adaptively adjust its default policy based on both the actual observed map and the \textit{imagined} unobserved map. 
Furthermore, the adaptive nature of the bounded rationality framework enables the robot to manage unreliable or incorrect imaginations by selectively sampling a few trajectories in the vicinity of the default policy.
Our approach utilizes a diffusion model for map prediction and a sampling-based planning with B-spline trajectory optimization to generate the default policy. Extensive evaluations reveal that the context-generative policy outperforms the baseline methods in identifying and avoiding unseen obstacles. Additionally, real-world experiments conducted with the Crazyflie drones demonstrate the adaptability of our proposed method, even when acting in environments outside the domain of the training distribution.
\end{abstract}

\section{INTRODUCTION}
\label{sec:intro}
Robotic autonomy in the real-world demands efficient decision-making despite constraints on sensing and computational capabilities. 
For instance, when navigating through a cluttered room, a robot cannot see through obstacles or behind walls, leading to a partial and potentially inaccurate understanding of its surroundings. 
Hence, the decision-making process operates within the constraints of this limited information, often leading to suboptimal choices such as taking inefficient paths that may lead to dead-ends due to occlusion.
These limitations highlight a sharp disparity with biological agents like humans, who effortlessly excel in tasks that robots find challenging. Despite these constraints, humans possess the remarkable ability to quickly select among a multitude of options by effectively narrowing down the search space. Leveraging their prior knowledge, humans constrain the range of choices available to them, opting for a satisfactory and sufficient (termed as {\em satisficing}~\cite{simon1987satisficing}) solution rather than an exhaustive pursuit of the optimal one. This decision-making paradigm is recognized as a bounded-rational decision-making process. 
Several models have emerged to tackle the challenges of bounded rationality (BR) among these, information-theoretic bounded 
rationality~\cite{ortega2015information, Ortega_IUBR, kappen2012optimal} stands out as a comprehensive framework explicitly accounting for a robot's computational limitations by imposing constraints on the amount of information it can process when transitioning from a default policy. The default policy serves as a collection of favorable choices from which an intelligent agent derives a satisficing solution for a given task and hence the efficacy of this default policy is highly reliant on prior experiences and accumulated knowledge. Recent advancements in this field~\cite{MABRA_DARS2022}, have assumed a default policy in the form of a normal distribution with no prior task-specific knowledge. Subsequent research~\cite{informedBR} has improved upon this approach by introducing a task-conditioned default policy that incorporates some prior knowledge. However, this prior work still assumes complete knowledge of the environment and relies on pre-trained goal-conditioned policies for navigating to any goal location within that environment. In scenarios involving partially known environments, we need more informative default policies that consider not only the goal but also the partial observations.

Rather than relying on a static default policy, in this study, we argue that the belief regarding favorable choices should dynamically evolve as new information becomes available, motivating the development of a context-generative default policy.
We present a novel approach that leverages observed environmental information to predict the unobserved space to facilitate adaptive adjustments in the agent's prior beliefs, aligning with the evolving knowledge of the world. This prior belief referred to as `context-generative default policy' is guided by design principles outlined in~\cite{informedBR}. We apply this approach to address the navigation challenge in unknown environments, achieving enhanced efficiency, as shown by the results of extensive simulation and real-world experiments. 

\section{RELATED WORK}
\label{sec:related_work}
The rationality assumption in the decision making process demands the evaluation of numerous decision options~\cite{papadimitriou1986intractable, dean1986intractability}.
One approach to mitigate this problem, as discussed in prior work~\cite{Zilberstein2011, zilberstein2011metareasoning, russell1991principles, boutilier1999decision}, is through meta-reasoning, wherein a robot examines the costs associated with selecting a choice by finding a balance between expected reward and the associated computational cost to evaluate the choice. However, this approach does not fundamentally reduce the number of choices, as they still optimize a transformed cost function under the assumption of rationality.
The concept of bounded rationality~\cite{russell1991principles, camerer1998bounded, selten1990bounded, zilberstein2011metareasoning, br-control} addresses this challenge by reducing the number of options evaluated based on prior task knowledge. Among various modeling approaches, the information-theoretic approach is particularly well-suited to robotics applications~\cite{ortega2015information, Ortega_IUBR}. This approach defines the decision problem as the minimization of divergence between the agent's default policy and the optimal policy~\cite{genewein2015bounded, ortega2013thermodynamics}, contending that a more informed default policy can lead to improved performance while evaluating a smaller set of choices.
However, prior works have primarily emphasized either uninformative default policies~\cite{MABRA_DARS2022} or goal-conditioned informed policies tailored to specific environments~\cite{informedBR}, which may prove insufficient when confronted with novel environments. We argue that the default policy should account for both the task and environmental factors to generate more informative candidate choices. Our approach achieves this by integrating predictions about environmental characteristics into the information-theoretic framework. 

Traditional methods in robotics~\cite{fox1997dynamic, quinlan1993elastic, williams2017model} usually overlook unobserved areas during the planning. While these methods can rapidly identify feasible trajectories, they often encounter difficulties in escaping local minima.
This challenge is tackled by progressively constructing a map during navigation~\cite{yamauchi1997frontier, pagac1998evidential, meyer2003map} but results in unnecessary exploration of the environment. 
Recent works involve using map prediction techniques to enhance the navigation efficiency as outlined in~\cite{map-predict-plan, katyal2021high, wang2021learning}. These methods introduce various approaches to map prediction and employ existing planners for motion control by assigning a lower cost to the predicted map. However, their performance is limited by the reliability of map prediction methods. Our proposed method adopts a similar approach to finding a default policy. However, unlike existing methods, it does not directly execute actions from the default policy. Instead, it evaluates a finite set of trajectories sampled from the default policy to effectively mitigate the unreliability of prediction methods.

\section{Method}
\label{sec:method}
We consider the problem of generating trajectories for a robot in an unknown environment represented by $e \in \mathcal{E}$, where $e$ can be any environment representation, i.e., occupancy or semantics; $\mathcal{E}$ is the set of all possible environments. Let $o_t$ be the area observed by the robot at given time $t$ in the given environment $e$. We define the context $c_t := \bigcup_{t=1}^{T} o_t$ where $c_t \subseteq e$. 
The goal of the robot is to find the trajectory $\tau $ that maximizes an expected reward given by 
\begin{equation}\label{eq:reward}
    J(\tau, s_t, g, c_t) = \sum_{k=t}^{t+H} R(s_t, a_t, g, e), 
\end{equation}
where $\tau = \{s_{t+1}, a_{t+1}, s_{t+2}, a_{t+2}, ..., s_{t+H-1}, a_{t+H-1}\}$, $s$ is the state, $a$ is the action, $g$ is the goal, $H$ is the planning horizon and $c$ is the context (e.g., a partially observed occupancy map).
We consider a deterministic dynamics $s_{t+1} = f(s_t, a_t)$. We interchangeably use the terms \(\tau\) and $a_t$ since we can iteratively apply action sequences to generate the trajectory \(\tau\).

Note that, diverging from the planning objective established in the prior work~\cite{informedBR}, the utility function $J$ now incorporates the contextual variable $c_t$ derived from the robot's operating environment $e$. Therefore, in this section, we first discuss the adjustments made in the problem formulation to integrate contextual variable and then introduce the concept of a context-generative default policy. Following that, we delve into the Bounded-Rational Policy Search Algorithm, which leverages the context-generative default policy for informed decision-making.

\subsection{Problem Modeling}
In principle, determining the optimal trajectory typically involves exhaustively evaluating every potential trajectory $\mathcal{T}$ within the state space when the environment $e$ is entirely known. However, such an exhaustive approach is often infeasible due to computational constraints inherent in robotic systems. Information-theoretic bounded rationality (ITBR) addresses this issue more realistically by explicitly accounting for agents' computational constraints in balancing computational resources against the number of evaluated trajectories~\cite{Ortega_IUBR}. 
This is achieved through a constraint on maximum expected utility by quantifying the amount of information the agent can process to transition from a default policy $Q$ to the optimal policy and formulated using KL-divergence as 
\begin{equation}\label{eq:constrained-value-fn}
\begin{aligned}
    \tau^{*}_t = \argmax_{\tau \sim Q}\{J(\tau, s_t, g, c_t)\}
    \textrm{, s. t. } D_{KL}(\pi_t || Q) \leq K,
\end{aligned}
\end{equation}
where $\tau$ is the trajectory sampled from the default policy, $\pi_t$ is the current policy of the robot, $D_{KL}$ is the KL divergence between the two stochastic policies, and $K$ is a constant denoting the amount of information an agent can deviate from the default policy. 
The default policy characterizes the robot's expected behavior prior to starting the navigation task. In bounded rational settings, a robot aims to find a {\em satisficing} trajectory within the proximity of the default policy. 
Using Lagrange multiplier, Eq.~\eqref{eq:constrained-value-fn} can further be reduced as:
\begin{equation}
\label{eq:itbr}
\begin{aligned}
    \tau^{*}_t = \argmax_{\tau \sim Q}\{J(\tau, s_t, g, c_t)\ - \Gamma(\beta, Q)\},
\end{aligned}
\end{equation}
where $\Gamma(\beta, Q) = \frac{1}{\beta} D_{KL}(\pi_t || Q)$, and $\beta > 0$ indicates the {\em rationality level}. 
Eq.~\eqref{eq:itbr} offers an enhanced capacity for modulating trade-off dynamics through the rationality parameter $\beta$ and the default policy $Q$.
In practice, the rationality level $\beta$ is linked to the computational capacity of the robot's hardware. Enhanced computational power allows for a more exhaustive evaluation of trajectories prior to decision-making, resulting in more rational decision outcomes. Since hardware flexibility is typically limited in robots, the degree of rationality predominantly relies on the informativeness of choices derived from the default policy. 
In this work, we introduce a novel perspective by associating the degree of rationality with the completeness of the robot's perception. Specifically, a greater range of perception enables evaluating trajectories more rationally, thereby facilitating escaping from potential local optima. Therefore, we propose a dynamic default policy that relies on the current state $s_t$, goal $g$, and context $c_t$, defined as: $Q_t(a_t \,|\, s_t, g, c_t)$
where $Q_t$ denotes the default policy and $a_t$ represents an action at time $t$. In this work, we treat the $\beta$ as a constant and aim to find a more informative default policy to improve robot's behavior.

Formally, the robot navigation under bounded-rational decision process is defined as a tuple ($\beta, \mathcal{T}_{Q_t}, Q_t, J$), where $\mathcal{T}_{Q_t} = \{\tau \sim Q_t \,|\, \text{Distance}(\tau, Q_t) \leq \epsilon\}$ represents the finite set of trajectories that are within a neighborhood (defined by $\epsilon$) of the default policy $Q_t$. 
The goal of the robot is to find a {\em satisificing} trajectory $\tau^* \sim \mathcal{T}_{Q_t} \subset \mathcal{T}$  that maximizes the following equation, derived from Eq.~\eqref{eq:itbr} following the prior work~\cite{informedBR, MABRA_DARS2022}:
\begin{equation}
\label{eq:cg-itbr}
\tau^{*}_t = \argmax_{\tau \sim \mathcal{T}_{Q_t}}\{-D_{KL}(\pi_t || \phi_t)\},
\end{equation}
where $\phi_t(a_t | s_t, g, c_t) \propto Q_t(a_t\,|\,s_t, g, c_t)e^{\beta J(\tau, s_t, g, c_t)}$.

\subsection{Context-Generative Default Policy}
In the ITBR framework, the quality of solutions is significantly influenced by the default policy choice. 
The default policy guides which part of the state space to search for the bounded-optimal policy.
Previous work leverages goal-conditioned reinforcement learning to compute an informative default policy. 
When the robot is deployed in a similar environment with a \textit{known} map, with the goal-conditioned informed policy, the agent finds a better policy using a small number of trajectory evaluations than the uniform (uninformed) one. 
However, when the environment map is not known a-prior, this informed policy can no longer provide correct guidance for the policy search.
Real-world environments typically exhibit consistent patterns and structures. 
Such regularities suggest that the unobserved segments of the environment can be extrapolated from the observed regions.
Conditioning the informative default policy on the unobserved regions allows the agent to utilize these environmental regularities to better guide the bounded-rational policy search to avoid potentially unobserved yet impassable regions.

\begin{figure}[t!] 
{
  \centering
  \subfigure  	
  {\includegraphics[width=0.41\textwidth]{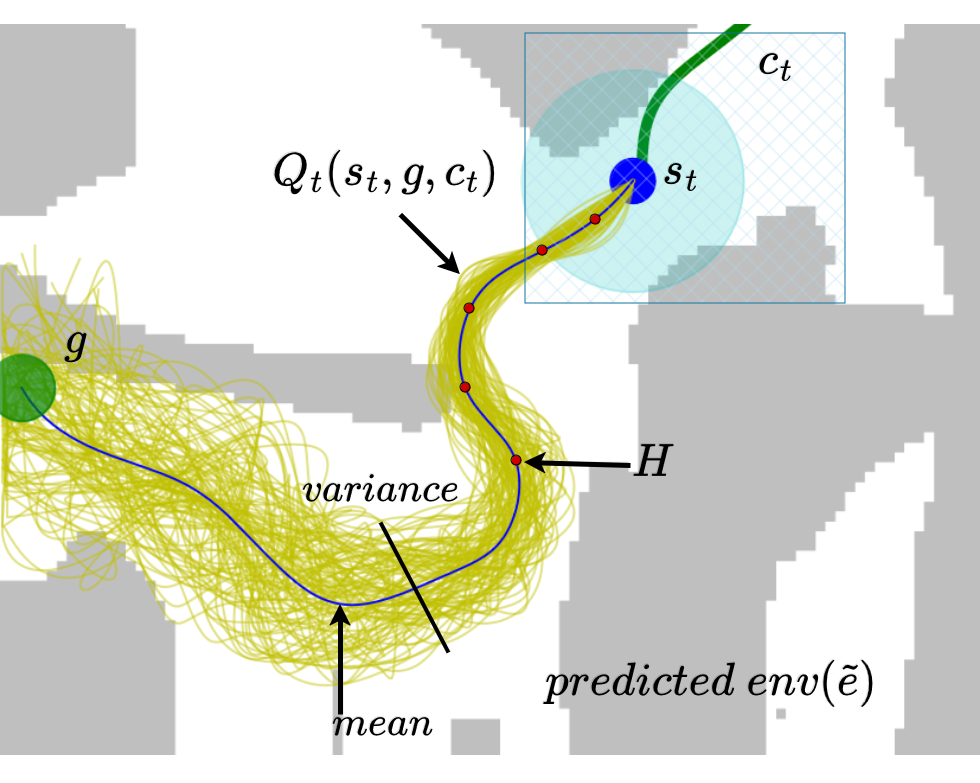}}\vspace{-10pt}
  \caption{Snapshot of the navigation task at time $t$, illustrating the default policy distribution. The predicted environment $\Tilde{e}$ is derived from the context $c_t$, indicated by the cross-hatched square. The yellow trajectories, sampled from $Q_t$, extend the horizon to the goal to enhance comprehension of the default policy but they can be truncated to any desired value during code implementation, as depicted by the red dots.
  \vspace{-10pt}
  }
\label{fig:qt} 
}
\end{figure}

Assume that we are given a model $\psi(\Tilde{e}|c_t; \theta)$ that facilitates the prediction of a complete map conditioned on the context $c_t$.
We use $\Tilde{e}$ to denote the map sampled from the model $\psi$ parameterized by $\theta$. 
To address the optimization problem outlined in~\eqref{eq:cg-itbr}, it is imperative to define $Q_t$. Hence, we adopt the design principles proposed in~\cite{informedBR} that states the designed default policy should be informative i.e., the sampled action sequences covers the high-utility regions, and it should be adaptive to changes. To ensure the informativeness, we use the model $\psi$ to anticipate the unseen environment and then use a sampling based-planner such as \textit{RRT$^*$} on predicted map $\Tilde{e}$ along with the \textit{B-spline} trajectory optimization to get the mean of $Q_t$. 
Finally, the path obtained from this planning process is truncated to the planning horizon $H$. We define the default policy as:
\begin{equation}
\label{eq:default-policy-def}
Q_t(\tau\,|\,\bar{p}, \Sigma) = \frac{1}{N} \exp\left(-\frac{1}{2} (\tau - \bar{p})^T \Sigma^{-1} (\tau - \bar{p})\right)
\end{equation}
where \( Q_t(\tau \,|\,\bar{p}, \Sigma) = Q_t(\tau\,|\,\gamma(s_t, g, \psi(\Tilde{e}|c; \theta)),\Sigma)\) with mean trajectory \( \bar{p} \) obtained from the path planning heuristics \(\gamma\) along which the policy is centered, the covariance matrix \( \Sigma \), \(N = \sqrt{(2\pi)^{H-1}|\Sigma|}\) is the normalization constant and \( \tau \) is the trajectory obtained by rolling out the sampled actions from the distribution.
Here, $\Sigma$ remains a tunable parameter, initially set to a low value at $s_t$ and gradually increased until $H$ is reached as shown in Fig.~\ref{fig:qt}.
The utility of the trajectories sampled from $Q_t$ depends on the accuracy of model $\psi$. If the predicted map significantly deviates from reality, it can diminish the informativeness of $Q_t$. Consequently, addressing inaccuracies in the map prediction becomes necessary to avoid such scenarios.
\subsection{Handling Map Prediction Errors}
Initially, in scenarios where robots have not yet gathered sufficient environmental data, we observe that $\Tilde{e}$ often tends to be inaccurate. This inaccuracy arises due to insufficient ground truth map input for the map prediction model. This situation is analogous to when biological agents navigate through unfamiliar environment; their expectations regarding unobserved area are typically inaccurate.
However, intelligent agents possess the capability to make decisions even in the presence of highly uncertain or potentially erroneous data. This capacity is referred to in the literature as `bounded information,' and real-world intelligent agents adeptly handle these limitations, making decisions efficiently despite constraints on time or the availability of accurate information.
Addressing these complex issues is challenging, and there is a well-documented argument suggesting that humans employ simple heuristics to navigate such scenarios, as discussed in~\cite{gigerenzer1999simple, gigerenzer2009homo}.
Hence, we take the following measures:
\begin{itemize}
    \item We assume that the goal always lies within navigable space and that a feasible path to the goal exists. When $\gamma$ fails to find a viable path, we adapt it to seek the nearest feasible path to the goal by directly linking the last accessible node to the goal and assigning minimal rewards to unreliable predictions, while still considering them as free to navigate.
    \item We further evaluate a finite number of trajectories proportional to the rationality number $\beta$, sampled from the default policy to find more informative path.
\end{itemize}  
This approach acknowledges the evident inaccuracies in the map prediction and offers a practical means of managing inaccurate information within bounded rational robotic agents.

\subsection{Context-Generative Bounded-Rational Policy Search}
The proposed method combines the above-mentioned context-generative default policy with ITBR framework. 
We compute the utility function by replacing $e$ with $\Tilde{e}$ in Eq.~\eqref{eq:reward} assuming the predicted map $\Tilde{e}$ is equivalent to the ground truth map $e$ given as:
\begin{equation}\label{eq:new-reward}
    J(\tau, s_t, g, c_t) = \sum_{k=t}^{t+H} R(s_t, a_t, g, \psi(\Tilde{e}|c_t;\theta)), 
\end{equation}
Note that the Eq.~\eqref{eq:cg-itbr} obtained its maximum value when $D_{KL}(\pi_t || \phi_t) = 0$ as KL-divergent being non-negative i,e., $\pi_t = \phi_t$ leading to the optimal action as shown below:
\begin{equation}
    \label{eq:optimal-action}
    \pi^*(a_t\,|\,s_t, g, c_t) = \frac{1}{\lambda}Q_t(a_t)e^{\beta J(\tau, s_t, g, c_t)}
\end{equation}
where \(\lambda = \int Q_t(a_t) e^{\beta \sum_{k=t}^{t+H} R(s_t, a_t, g, \psi(\Tilde{e}|c_t;\theta))} \, da_t
\) is the normalization constant. Finally, the expected optimal action sequences is obtained by importance sampling given by:
\begin{equation}\label{eq:importance-sampling}
\begin{split}
    \mathbb{E}_{a^*_t \sim \pi^{*}_{t}}[a_t\,|\,s_t, g, c_t]  
           \approx \frac{\sum_{k=1}^{L} w(a_{t,k}) a_{t,k}}{\sum_{k=1}^{L} w(a_{t,k})}.
\end{split}
\end{equation}
where \( w(a_{t,k}) = e^{\beta \sum_{k=t}^{t+H} R(s_{t,k}, a_{t,k}, g, \psi(\Tilde{e}|c_t;\theta))}\) and $L$ is the number of samples to be evaluated.
In summery, the robot navigation involves sampling $L$ actions from a finite number of choices $\mathcal{T}_{Q_t}$ obtained from $Q_t$. Subsequently, each sequence is assigned a weight $w$ that reflects its significance according to the agents' utilities and level of rationality and the reward function is computed by rolling out future states along the action sequence until the planning horizon ends. Finally, the expected optimal action is obtained by Eq.~\eqref{eq:importance-sampling}. 


\begin{figure*}[t!] \vspace{-8pt}
  \centering
  \subfigure[Distance to Goal vs Time]
  	{\includegraphics[height=1.4in, width=2.2in]
    {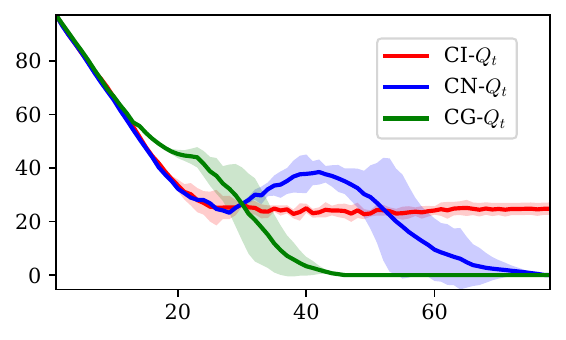}
         \label{fig:dist2goal1}}
  \subfigure[Explored Area vs Time]
  	{\includegraphics[height=1.4in, width=2.2in]
    {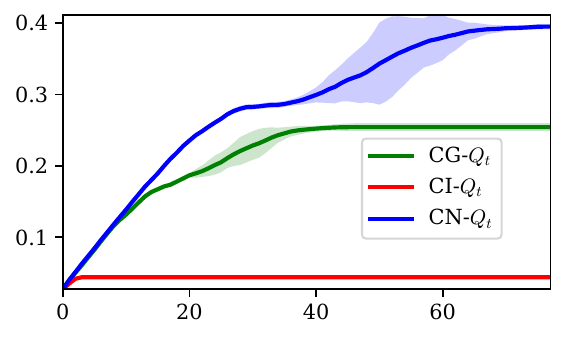}
         \label{fig:nav_eff1}}
  \subfigure[$M_{Acc}$ vs Time]  	
    {\includegraphics[height=1.4in, width=2.2in]{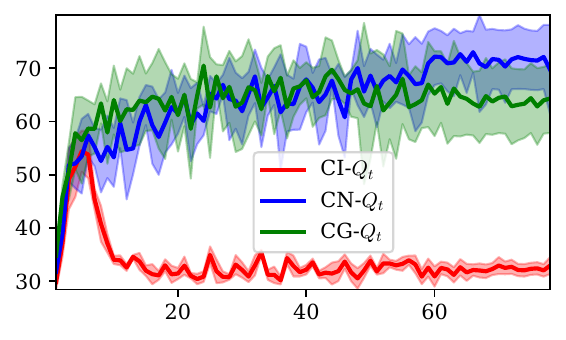}
         \label{fig:acc1}}
  \caption{Illustration of the performance evaluation with respect to navigation task. The X-axis indicates the time while the Y-axis indicates the distance to goal in (a), the explored area in (b), and map prediction accuracy in (c).
  \vspace{-10pt}
  }
\label{fig:A}  
\end{figure*}

\vspace{5pt}
\section{Simulation Experiments}
\label{sec:sim_exp}


\subsection{Simulation Setup and Performance Metrics}
We conducted two experiments to validate our approach. The first evaluates the robot's navigation abilities, focusing on its capacity to avoid challenging obstacles in a 2D environment. The second experiment analyzed the influence of initial context (landmark knowledge) on the performance.

\textbf{Map Prediction Model: }
We use the {\em nuScenes} dataset for training our map-prediction module, as documented in \cite{caesar2020nuscenes}. 
We convert the semantic images into occupancy maps by categorizing the classes into two groups: navigable and non-navigable. The resulting dataset is divided into two subsets: a training set comprising 28,008 images and a separate testing set consisting of 500 images. All the simulation testing environments are drawn from this pool of 500 images.
We use the SePaint model~\cite{chen2023sepaint} to predict the environment.

\textbf{Performance Metric: } We employ a set of robust metrics that includes {\em path length}, {\em navigation efficiency} ($N_{eff}$), and {\em map prediction accuracy} ($M_{Acc}$). The navigation efficiency ($N_{eff}$) is a critical measure, quantifying the change in distance to the goal relative to the explored area at a specific time step $t$ within the given environment. Higher $N_{eff}$ values signify more efficient navigation, indicating that the agent covers a shorter spatial extent while making substantial progress toward its goal. This not only highlights the agent's ability to minimize detours but also emphasise its energy-saving potential for future tasks. Map prediction accuracy ($M_{Acc}$) is defined as $M_{Acc} = (n_{tp}+n_{tn})/(n_{tp}+n_{tn}+n_{fp}+n_{fn})$, where $n_{tp}, n_{tn}, n_{fp}$, and $n_{fn}$ represent true positives, true negatives, false positives, and false negatives, respectively.

\textbf{Constant Parameters in ITBR: } We employ consistent settings for key parameters. The agents' transition functions are determined by a deterministic single integrator model, where their speed is bounded between $0 m/s$ (minimum speed) and $1 m/s$ (maximum speed). 
Our one-step reward function is thoughtfully designed to discourage collisions and substantial deviations from the goal. To ensure consistency in our experiments, we maintain a fixed number of sampled trajectories at 100. Additionally, we keep the rationality level ($\beta=0.04$) constant throughout our simulations.

\textbf{Baselines: }
We use two baselines to compare with the proposed method. The first baseline disregards the context acquired by the agent, treating the unknown areas as free to navigate and generating a default policy similar to~\cite{informedBR}. We refer to this baseline as the {\em ``Context-Ignorant Default Policy (CI-$Q_t$)"}. The second baseline uses the observed map but ignores its potential to predict the unknown map and only rely on the known map for generating the default policy, referred as {\em ``Context-Neutral Default Policy (CN-$Q_t$)"}. Our proposed method is referred as {\em ``Context-Generative Default Policy (CG-$Q_t$)"}. 
\begin{figure*}[t!] 
  \centering
  \subfigure[CG-$Q_t$ $(t_1)$]
  	{\includegraphics[height=1.0in, width=1.3in]
    {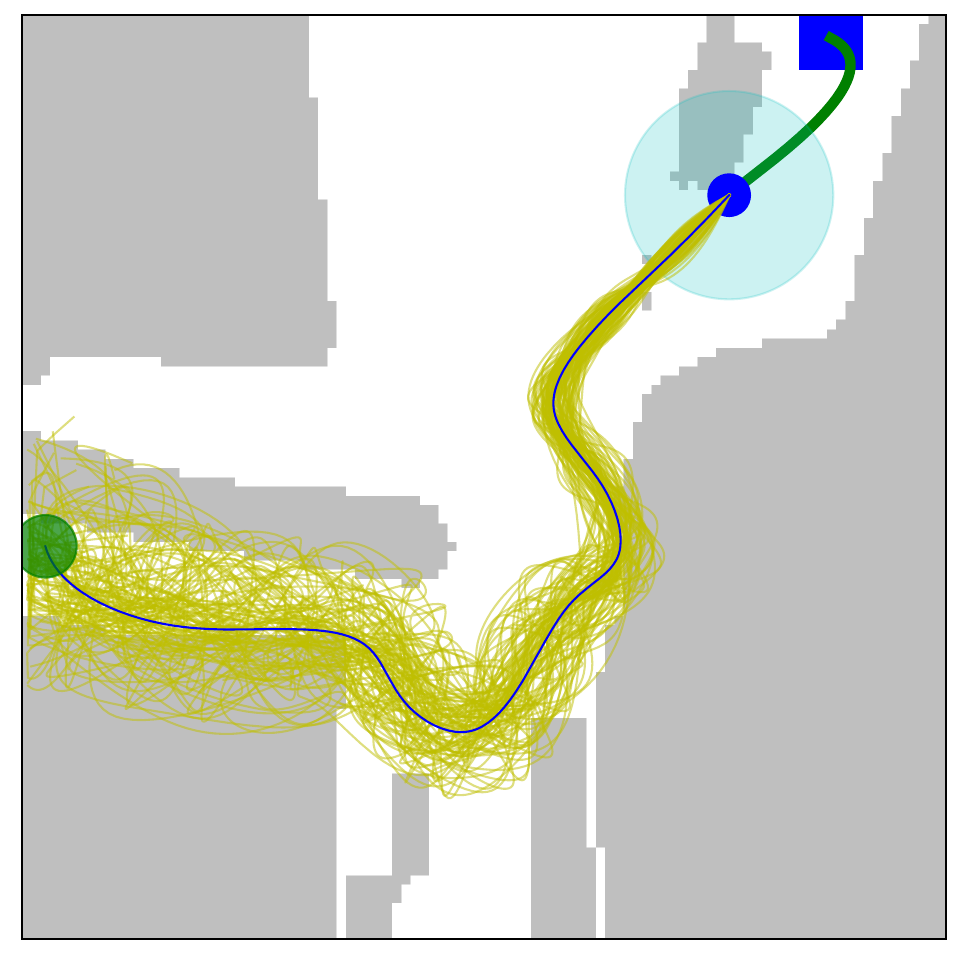}
         \label{fig:Inf_t1}}
  \subfigure[CG-$Q_t$ $(t_2)$]
  	{\includegraphics[height=1.0in, width=1.3in]
    {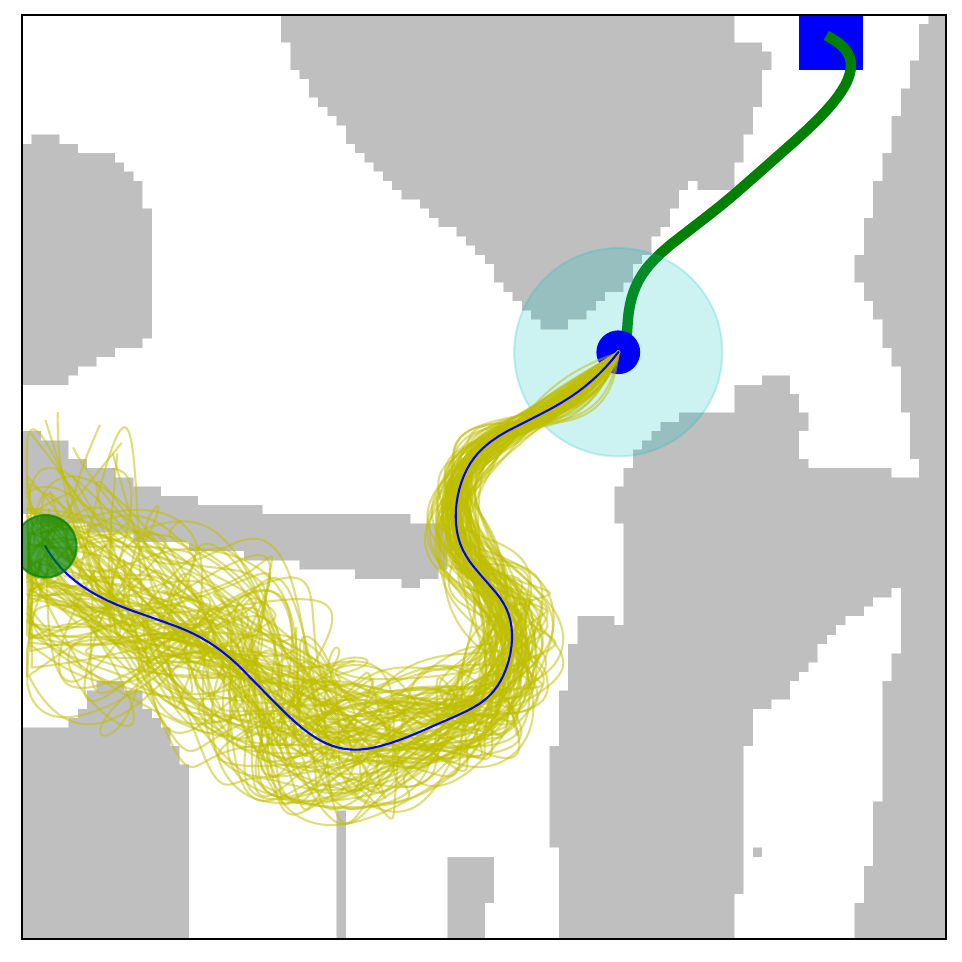}
         \label{fig:Inf_t2}}
  \subfigure[CG-$Q_t$ $(t_3)$]  	
    {\includegraphics[height=1.0in, width=1.3in]{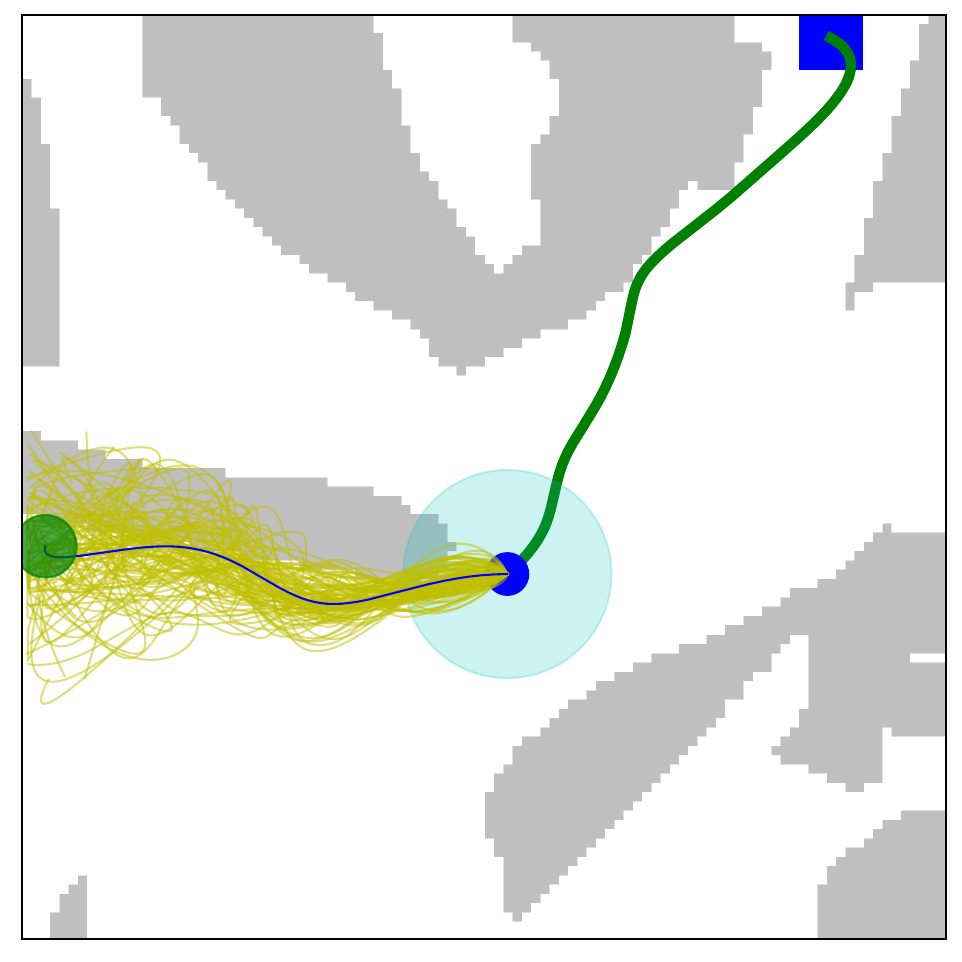}
         \label{fig:Inf_t3}}
  \subfigure[CG-$Q_t$ $(t_4)$]
  	{\includegraphics[height=1.0in, width=1.3in]
    {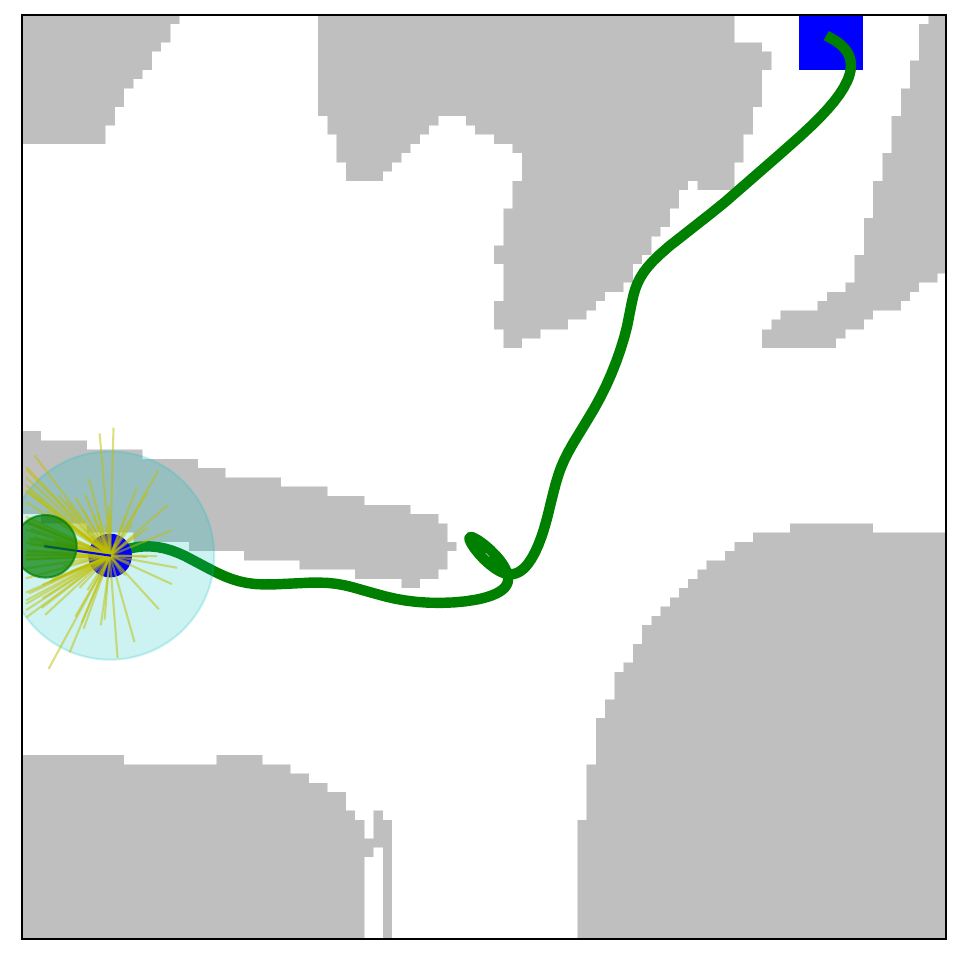}
         \label{fig:Inf_t3}}
  \subfigure[CG-$Q_t$ $(t_5)$]
  	{\includegraphics[height=1.0in, width=1.3in]
    {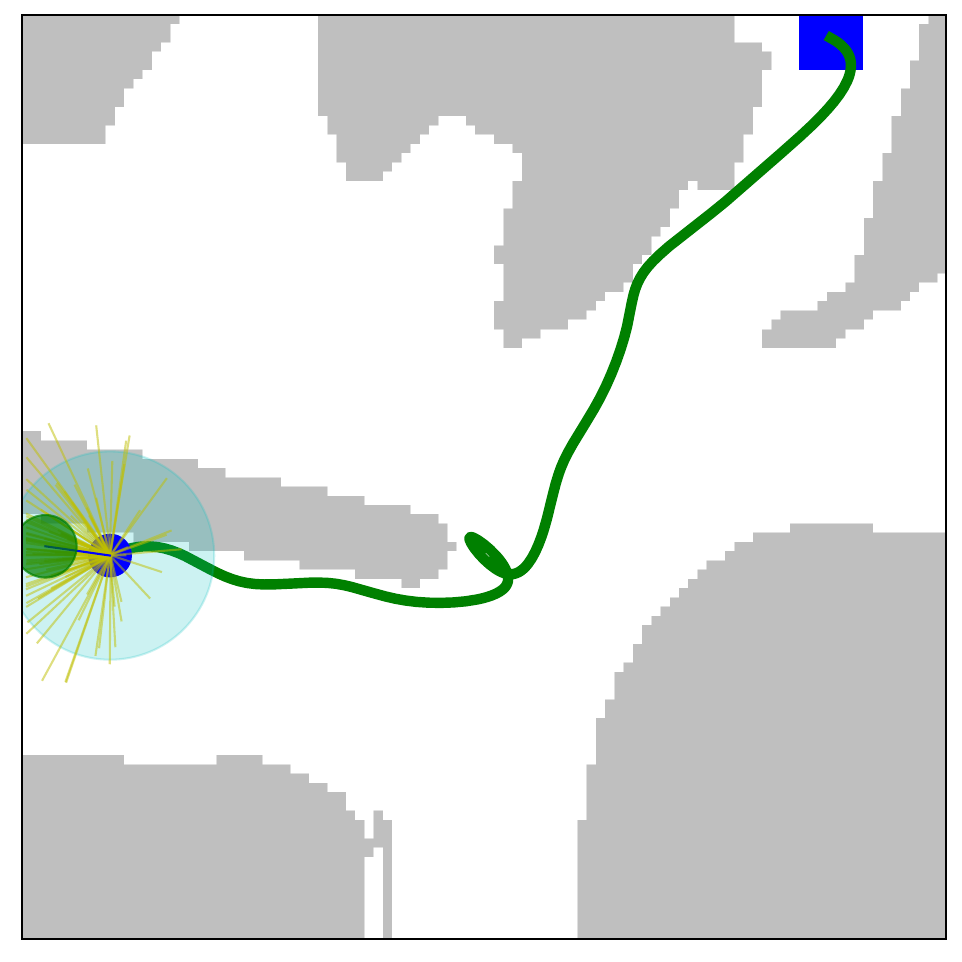}
         \label{fig:Inf_t5}}
  \subfigure[CN-$Q_t$ $(t_1)$]
  	{\includegraphics[height=1.0in, width=1.3in]
    {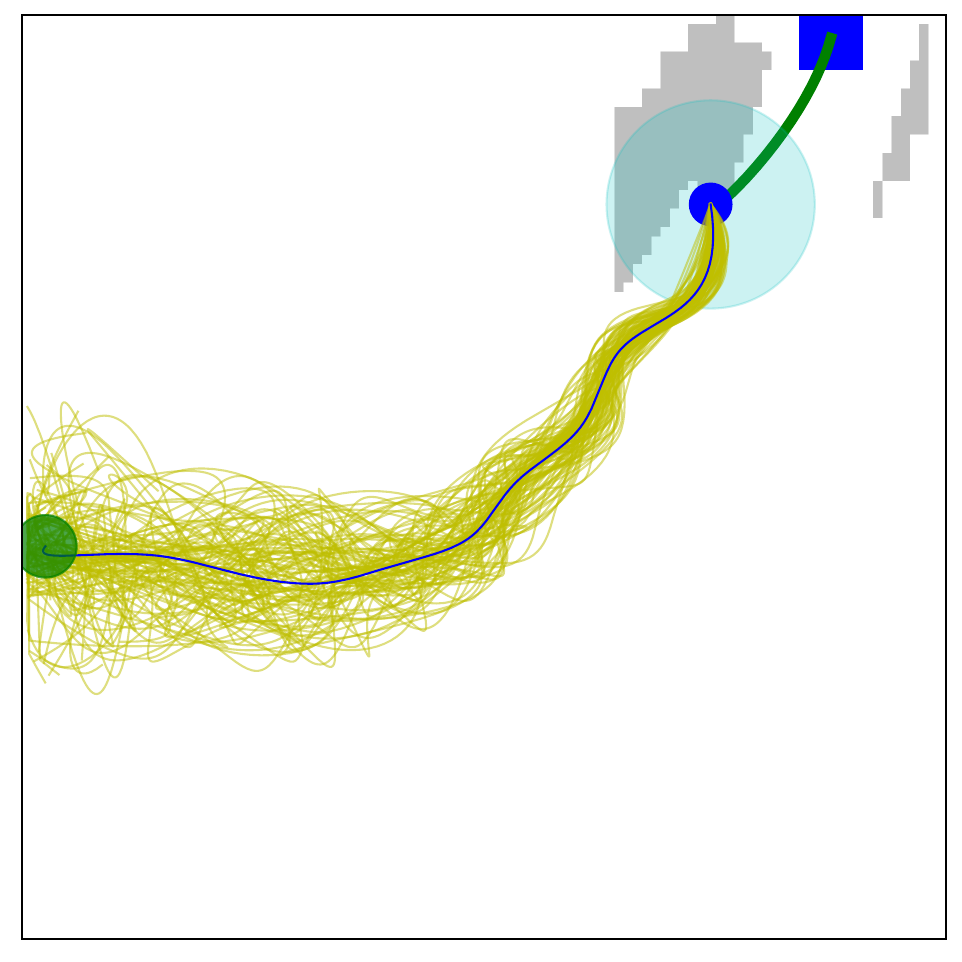}
         \label{fig:U_t1}}
  \subfigure[CN-$Q_t$ $(t_2)$]
  	{\includegraphics[height=1.0in, width=1.3in]
    {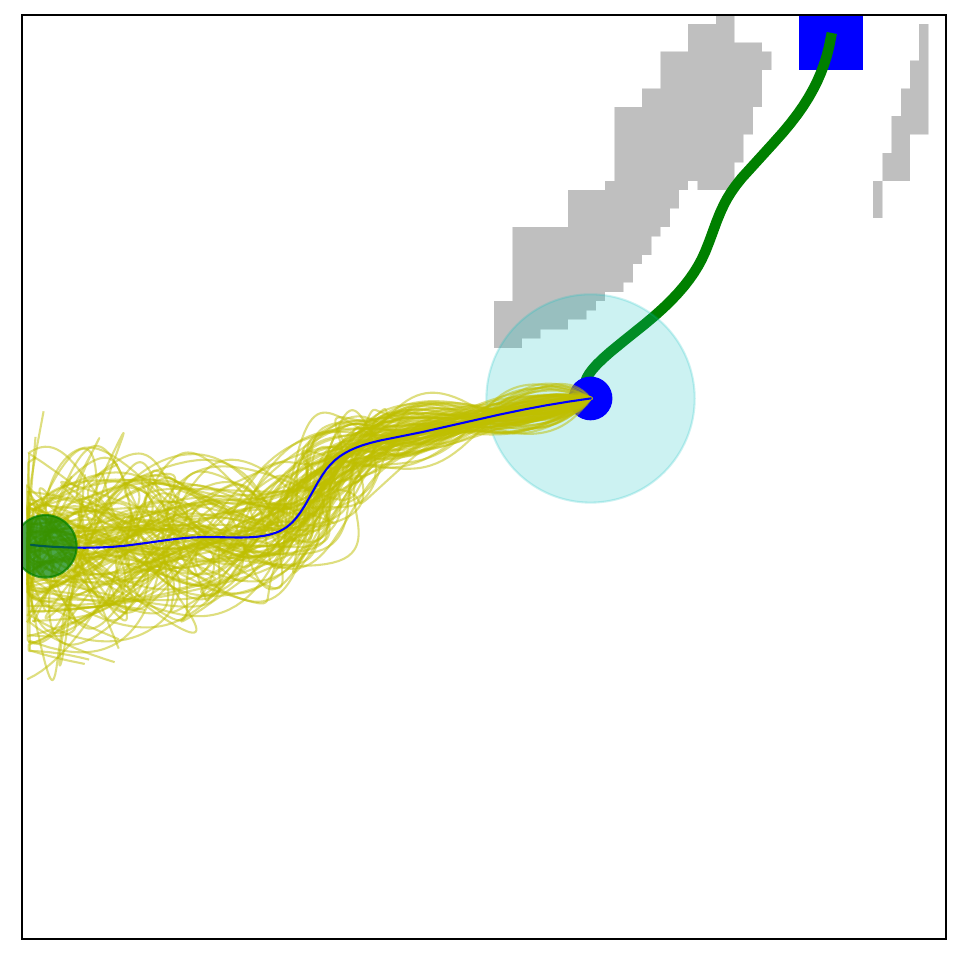}
         \label{fig:U_t2}}
  \subfigure[CN-$Q_t$ $(t_3)$]  	
    {\includegraphics[height=1.0in, width=1.3in]{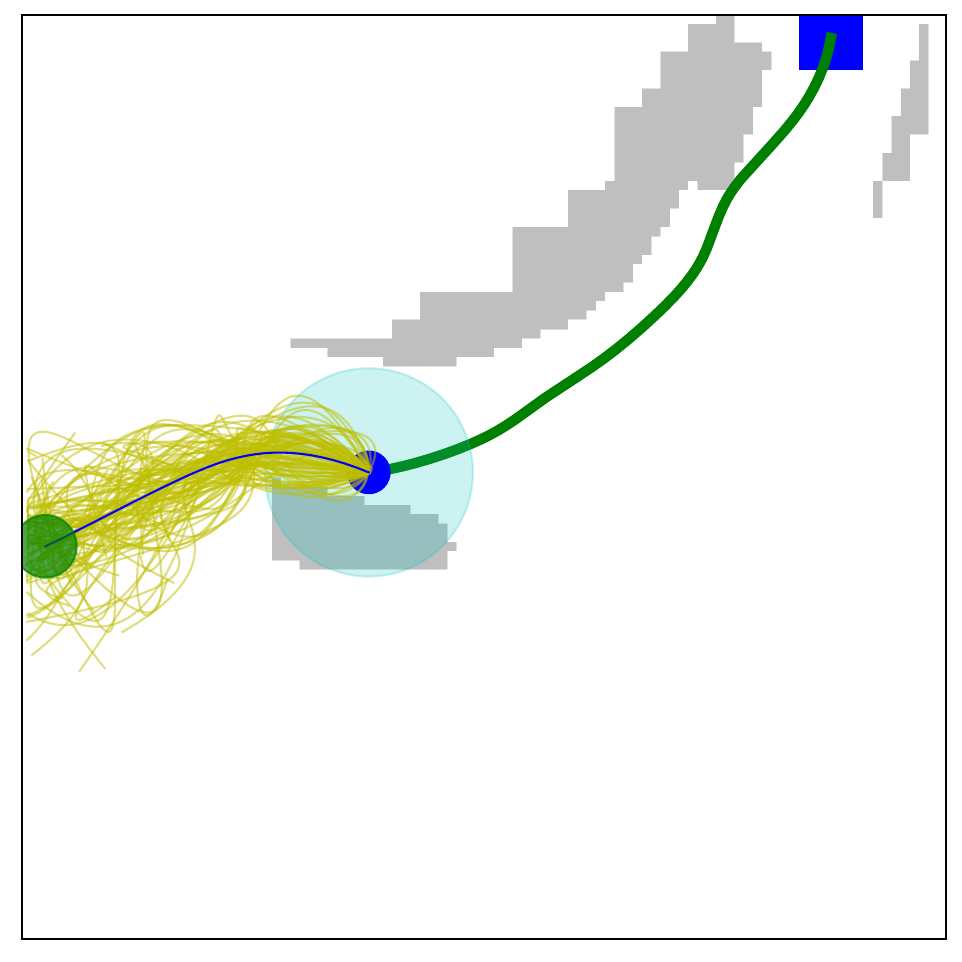}
         \label{fig:U_t3}}
  \subfigure[CN-$Q_t$ $(t_4)$]
  	{\includegraphics[height=1.0in, width=1.3in]
    {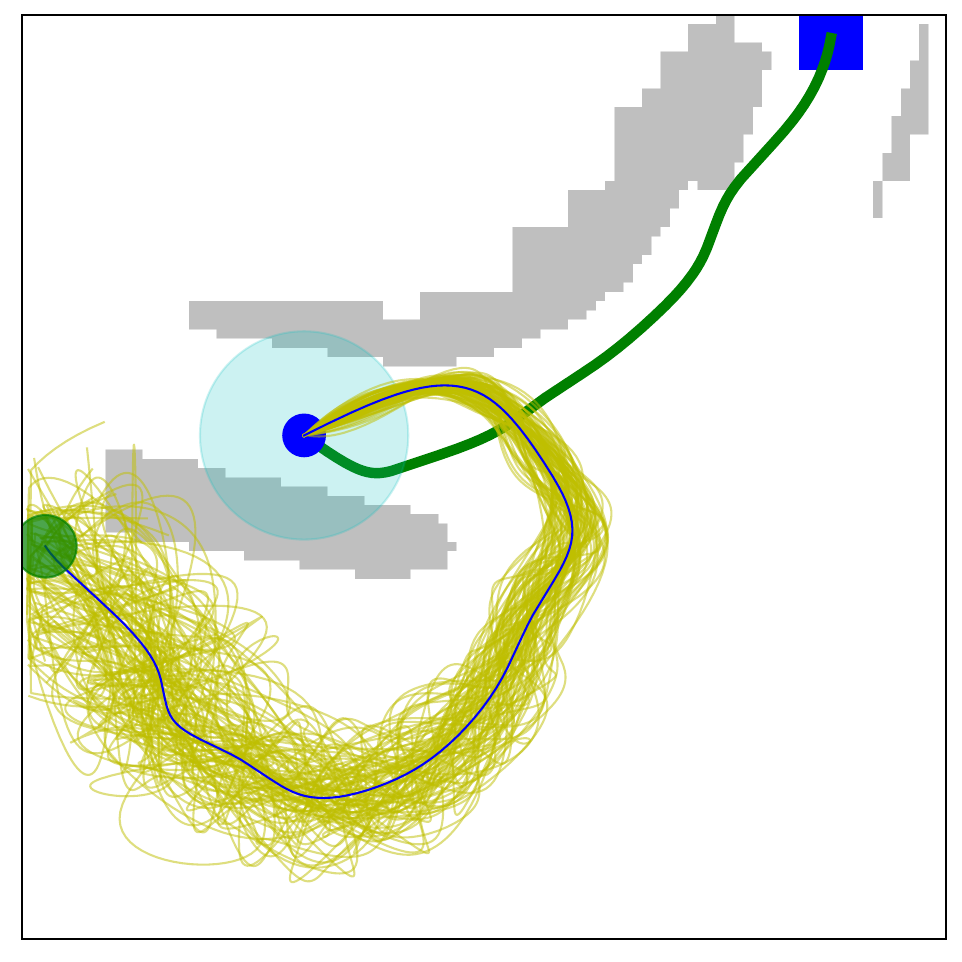}
         \label{fig:U_t3}}
  \subfigure[CN-$Q_t$ $(t_5)$]
  	{\includegraphics[height=1.0in, width=1.3in]
    {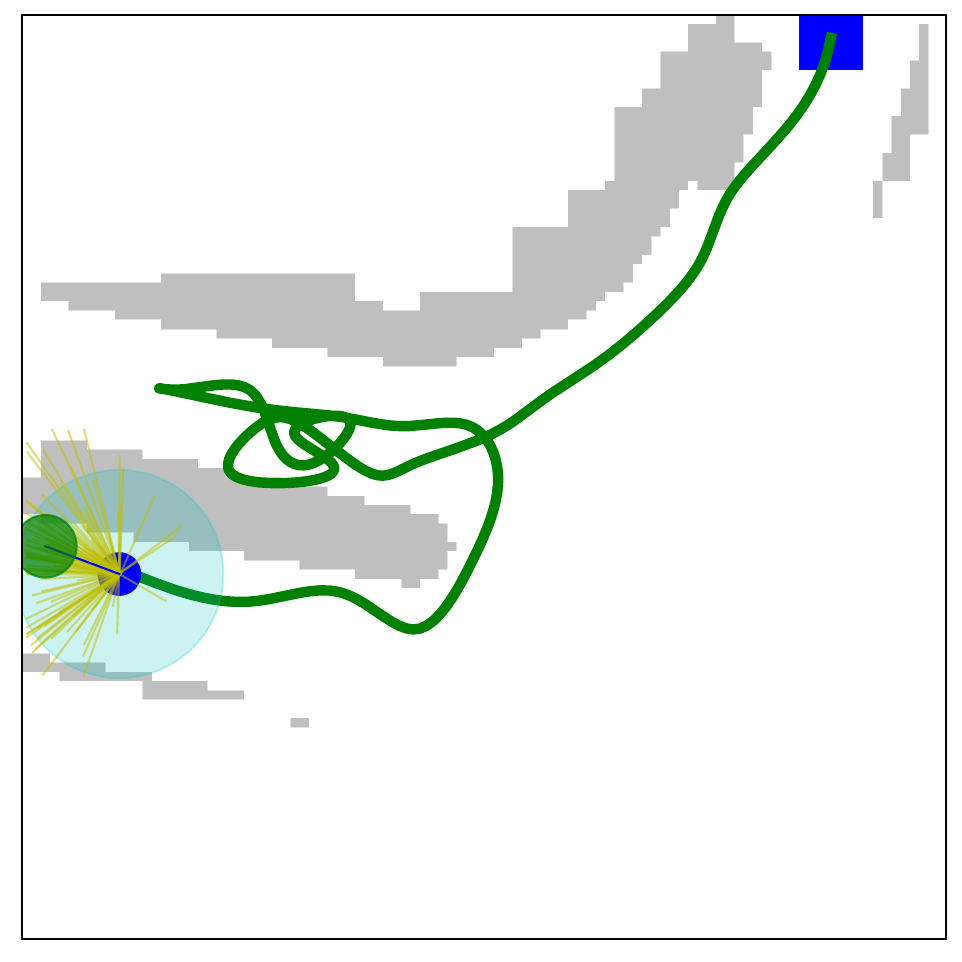}
         \label{fig:U_t5}}
  \subfigure[CI-$Q_t$ $(t_1)$]
  	{\includegraphics[height=1.0in, width=1.3in]
    {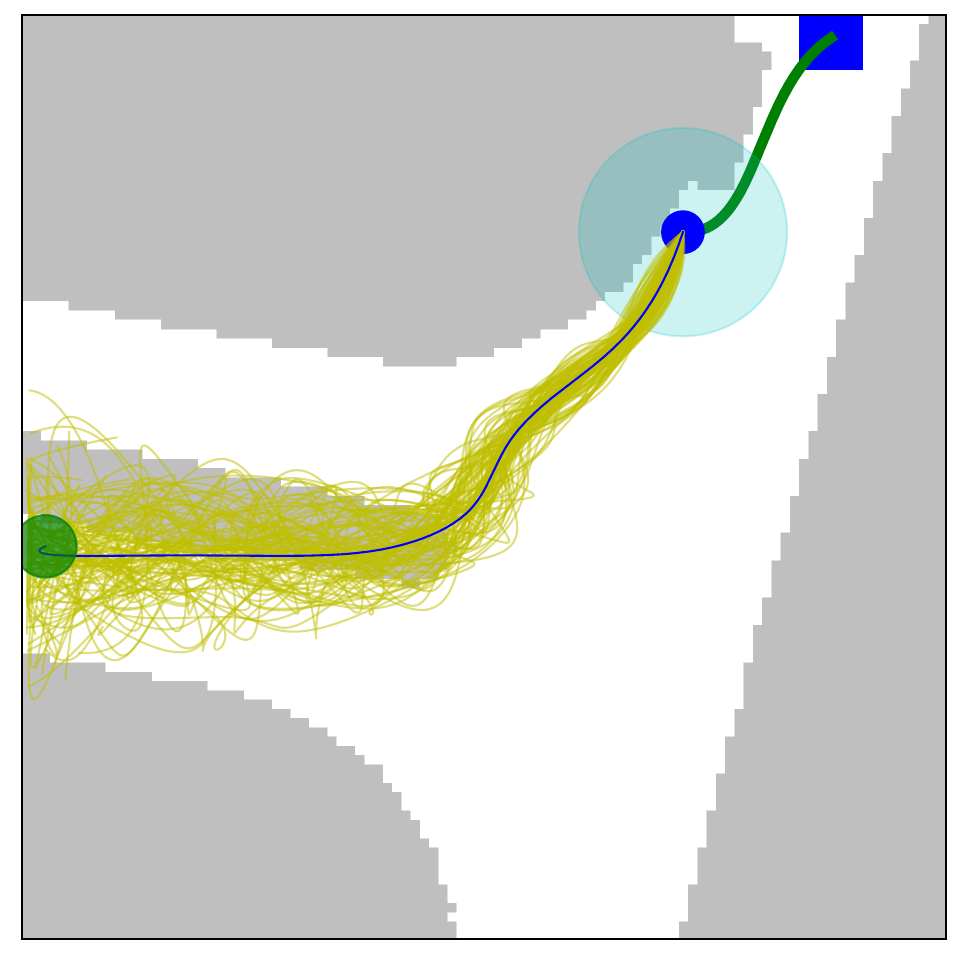}
         \label{fig:U_t1}}
  \subfigure[CI-$Q_t$ $(t_2)$]
  	{\includegraphics[height=1.0in, width=1.3in]
    {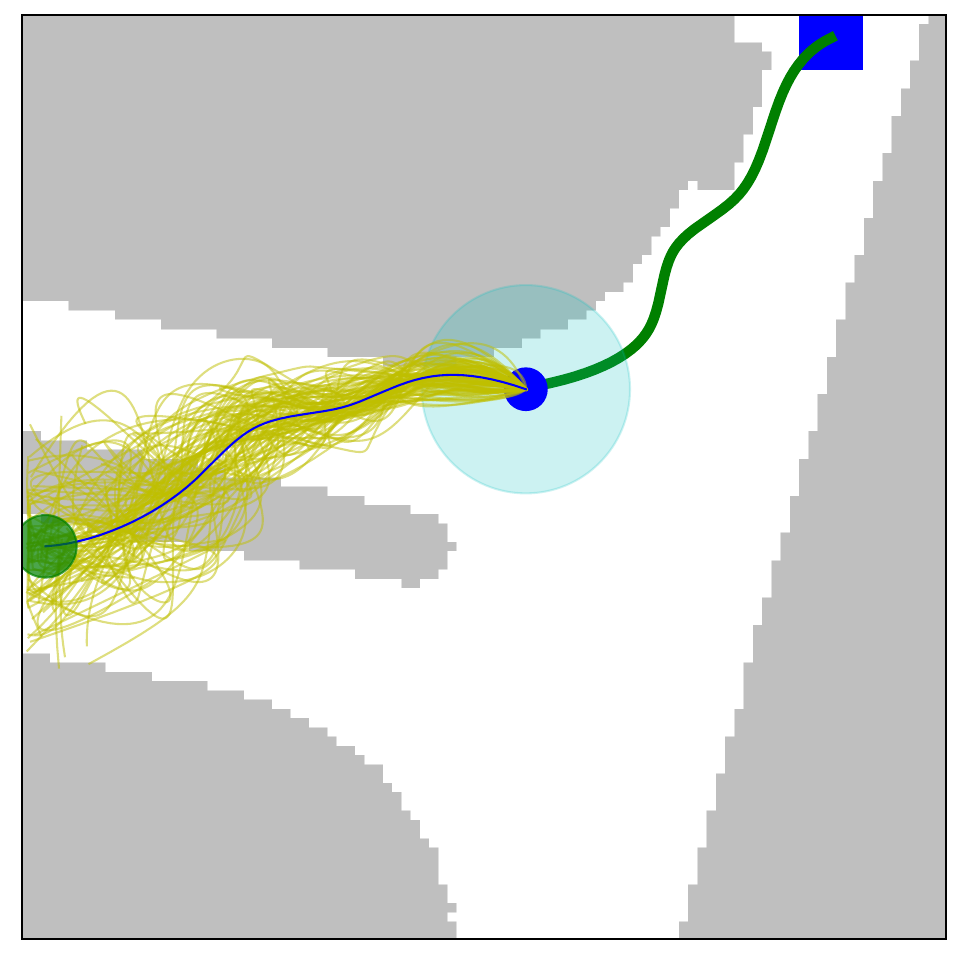}
         \label{fig:U_t2}}
  \subfigure[CI-$Q_t$ $(t_3)$]  	
    {\includegraphics[height=1.0in, width=1.3in]{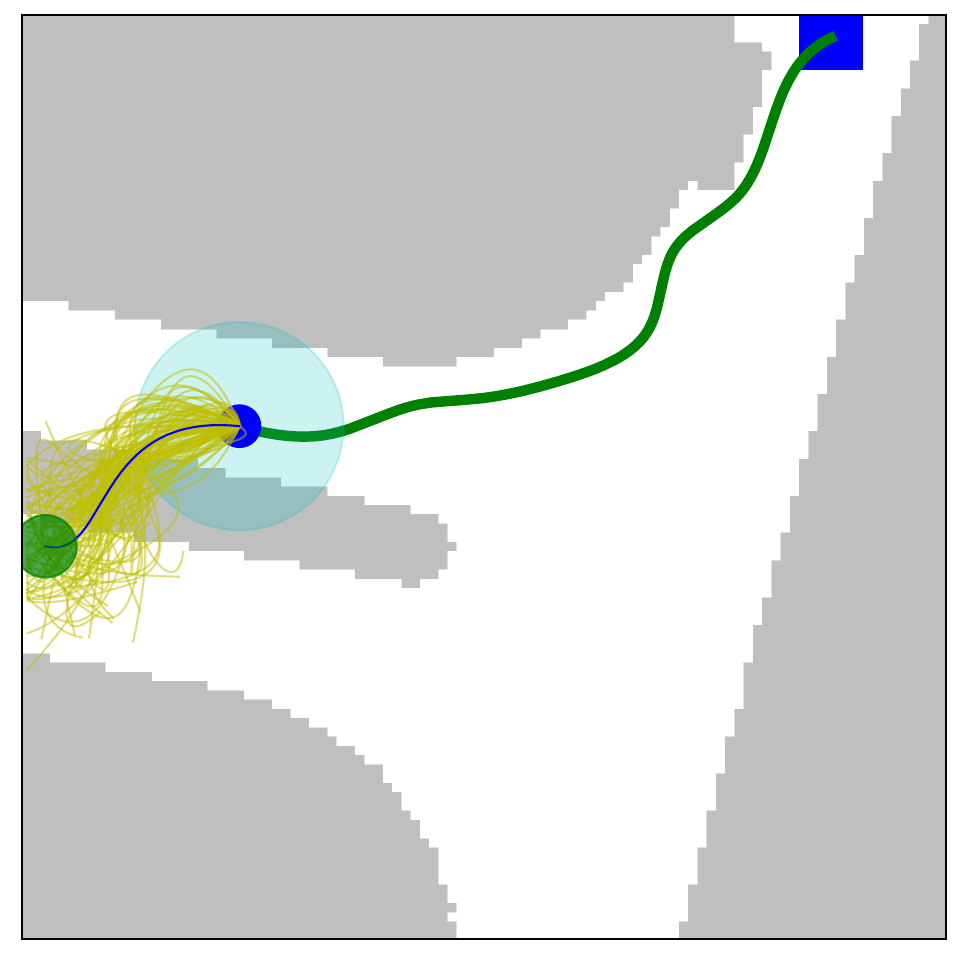}
         \label{fig:U_t3}}
  \subfigure[CI-$Q_t$ $(t_4)$]
  	{\includegraphics[height=1.0in, width=1.3in]
    {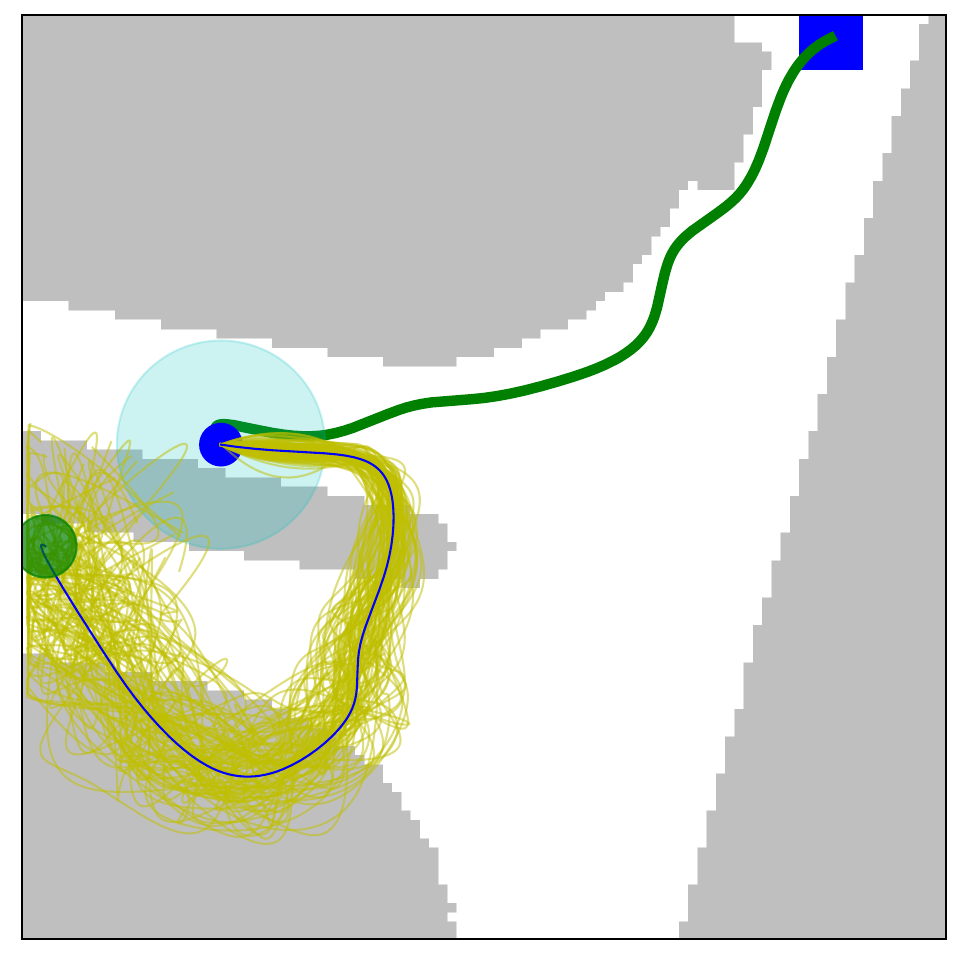}
         \label{fig:U_t3}}
  \subfigure[CI-$Q_t$ $(t_5)$]
  	{\includegraphics[height=1.0in, width=1.3in]
    {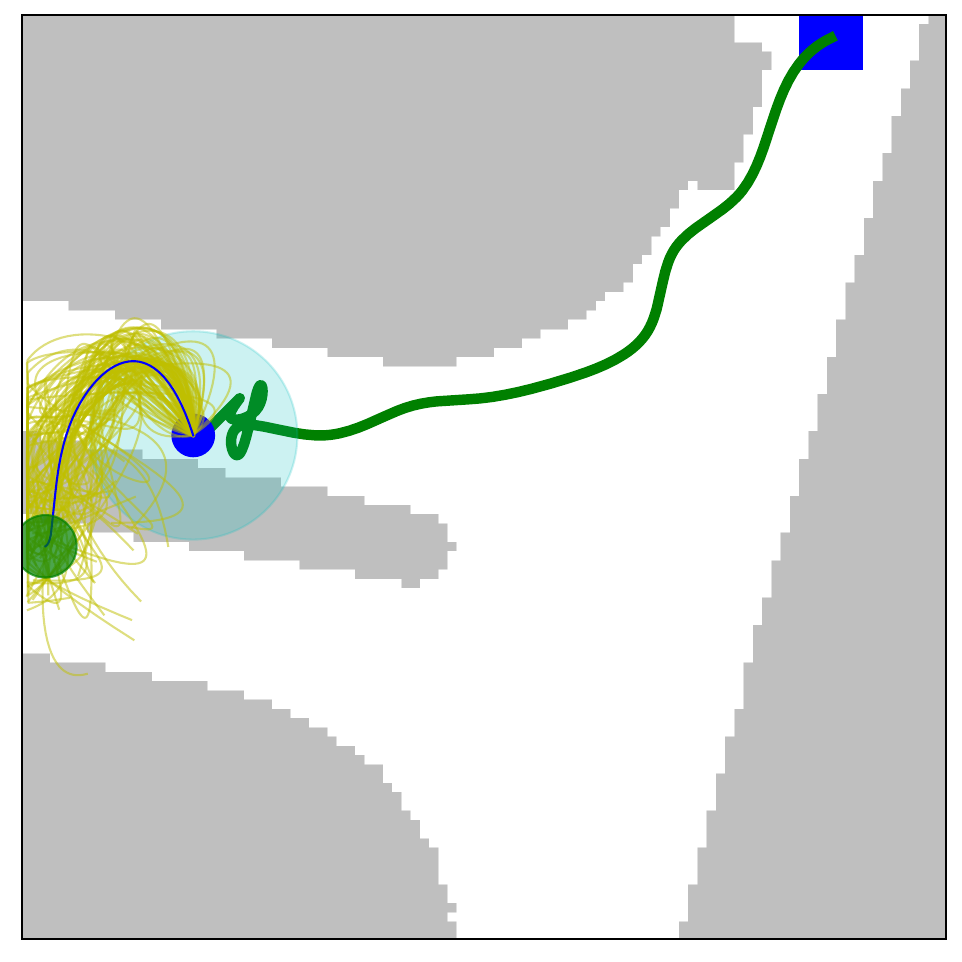}
         \label{fig:U_t5}}
  \caption{Illustration of a grid environment with the starting point (blue square), robot's current position (blue circle), and the sensor field of view (cyan circle). The path traveled by the robot is shown in green, the yellow paths represent the default distribution with predicted mean path. (a) to (e) showcase the performance of the proposed method on the predicted map. (f) to (j) illustrate the performance of the CN-$Q_t$ method on the observed map. (k) to (o) present the performance of the CI-$Q_t$ method, which only considers sensor's field of view, the paths are overlaid on the ground truth map. All snapshots were captured simultaneously from the starting point to facilitate direct comparison.
  \vspace{-12pt}
  }
\label{fig:path_comp}  
\end{figure*}
\subsection{Anticipating Obstacles: Beyond Dead Ends
}
We empirically demonstrate the remarkable anticipatory capabilities of our proposed method in navigating complex environments, specifically in scenarios involving challenging obstacles. To evaluate this, we conducted controlled experiments where we randomly sampled a map from the testing environment, deliberately selecting scenarios with complex obstacles, such as U-shaped barriers placed between the starting point and the goal.

As depicted in Fig.~\ref{fig:dist2goal1}, our proposed method and the CN-$Q_t$ approach both successfully completed the navigation task. In contrast, the CI-$Q_t$ approach encountered difficulties, evident from the consistent and prolonged tail in the curve, indicative of being trapped in local minima. Furthermore, while the CN-$Q_t$ approach ultimately completed the task, it briefly deviated from the optimal path, leading to an increase in distance to the goal - a clear indication of navigating into a dead end. Fig.~\ref{fig:path_comp} shows a detailed comparative illustration of this experiment, with simultaneous snapshots for reference. 
To evaluate the influence of acquired information during navigation on map prediction methods, we provide the explored area by all the methods at a given time (shown in Fig.~\ref{fig:nav_eff1}) as contextual information to the map prediction process and recorded their prediction accuracy. Our findings indicate the substantial impact of incorporating increased context on the map prediction accuracy as shown in Fig.~\ref{fig:acc1}. 
This experiment was repeated across 10 randomly selected maps, consistently yielding the same results. These outcomes establish our proposed method as a promising asset for addressing long-range navigation tasks, where ample context is acquired during the initial exploration, leading to improved prediction accuracy and offering a potential solution to real-world challenges in robotics.
\begin{figure*}[t!] \vspace{-8pt}
  \centering
  \subfigure[Distance to Goal vs Time]
  	{\includegraphics[height=1.43in, width=2.2in]
    {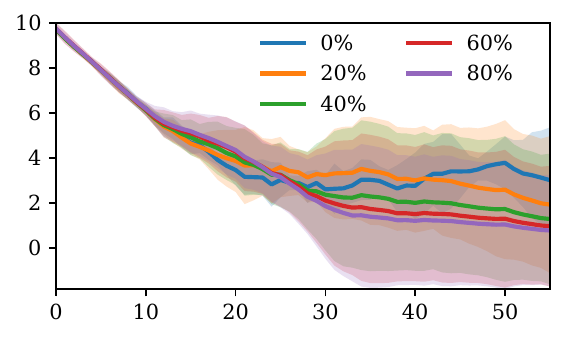}
         \label{fig:dist2goal2}}
  \subfigure[Navigation Efficiency vs Time]
  	{\includegraphics[height=1.4in, width=2.2in]
    {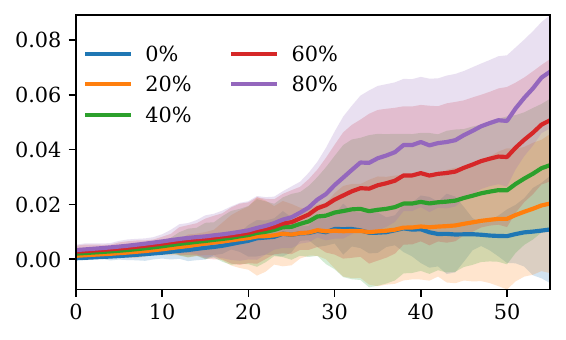}
         \label{fig:nav_eff2}}
  \subfigure[Pathlength vs Maps]  	
    {\includegraphics[height=1.4in, width=2.2in]{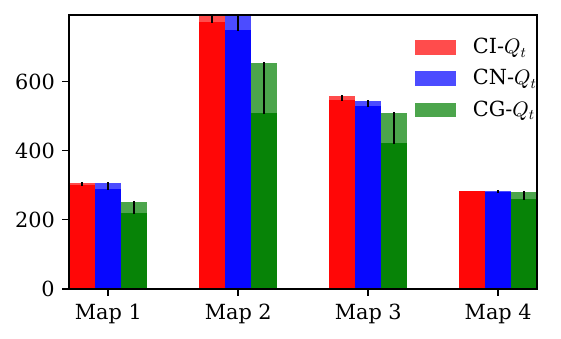}
         \label{fig:B3}}
  \subfigure[Map 1]
  	{\includegraphics[height=1.3in, width=1.65in]
    {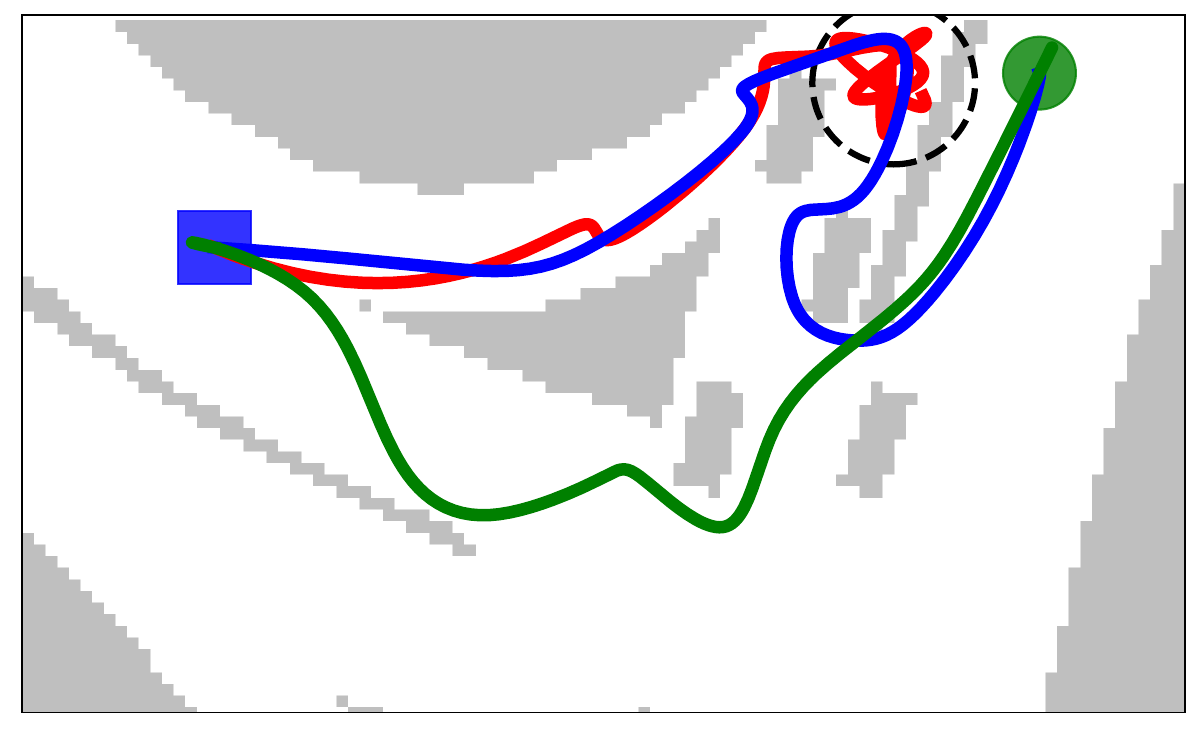}
         \label{fig:env0}}
  \subfigure[Map 2]
  	{\includegraphics[height=1.3in, width=1.65in]
    {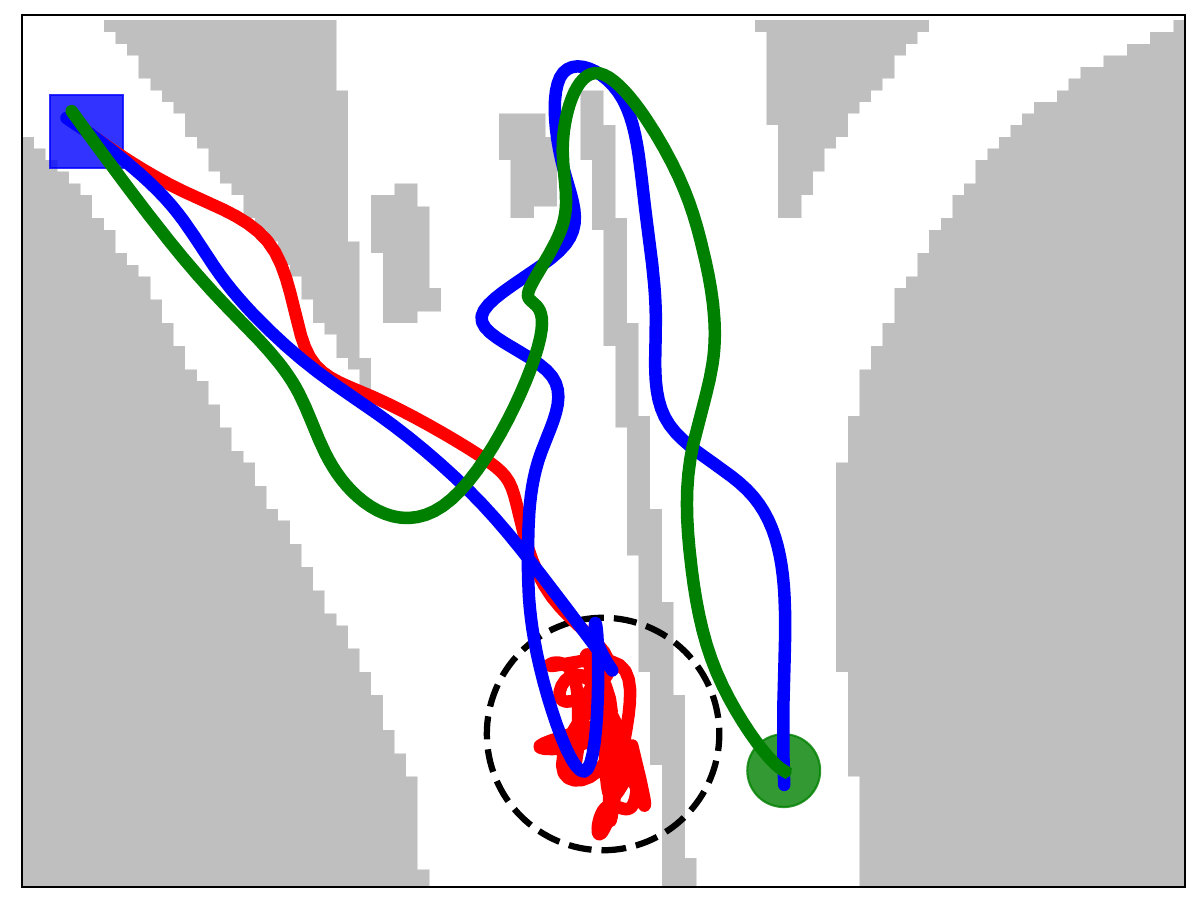}
         \label{fig:env1}}
  \subfigure[Map 3]  	
    {\includegraphics[height=1.3in, width=1.65in]
    {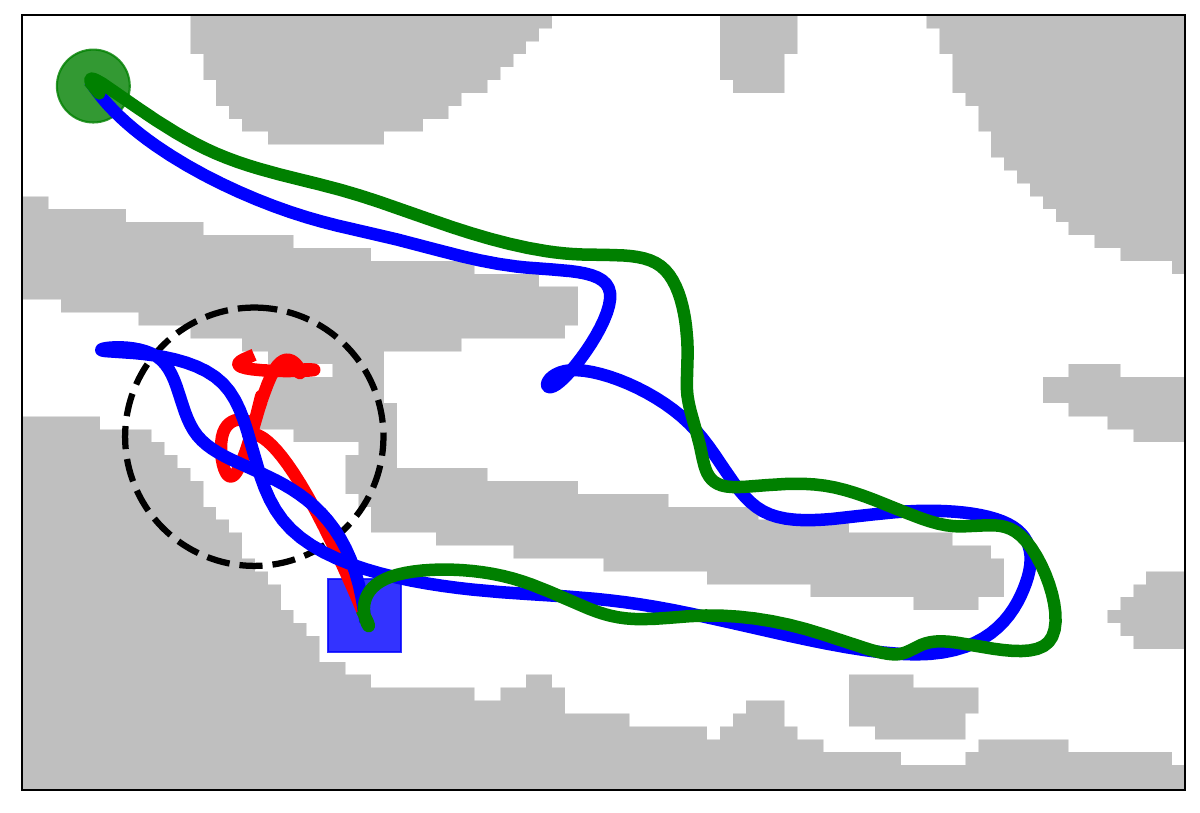}
         \label{fig:env2}}
  \subfigure[Map 4]
  	{\includegraphics[height=1.3in, width=1.65in]
    {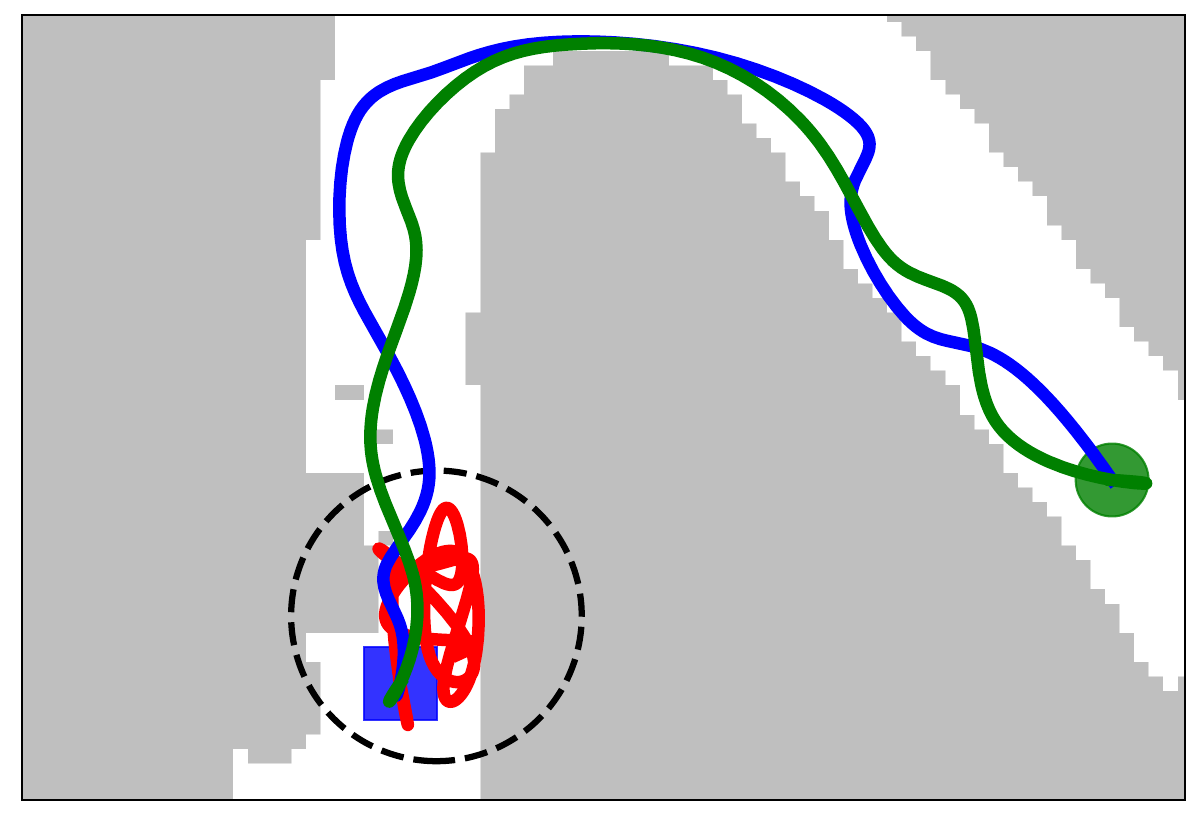}
         \label{fig:env3}}
  \caption{Illustration of the impact of providing initial context on the navigation task and the performance of baselines and proposed method in four different environments. (a) and (b) shows the performance of proposed method on increasing initial context where x-axis represents time. Y-axis represents the distance to goal in (a) and navigation efficiency in (b). (c) demonstrates the path length on y-axis for 4 different maps. Note that the vertical black line shows the difference in the improvement for path length when given more initial context. The path travelled by all the methods on the 4 maps are visualised in (d) to (g) in which the start position is represented by blue square and goal is shown by the green circle. The black dotted circles on the map highlights the area where baselines encounter difficulties. Red path represents CI-$Q_t$, blue path represents CN-$Q_t$ and green path represents CG-$Q_t$.
  \vspace{-12pt}
  }
\label{fig:multi_agent_exp}  
\end{figure*}
\begin{figure*}[t!]
    \centering
    \centering
  \subfigure[Initial Map]
  	{\includegraphics[height=1.2in, width=1.65in]
    {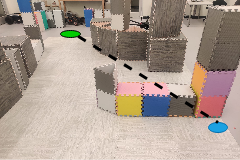}
         \label{fig:renv0}}
  \subfigure[CG-$Q_t$ Path]
  	{\includegraphics[height=1.2in, width=1.65in]
    {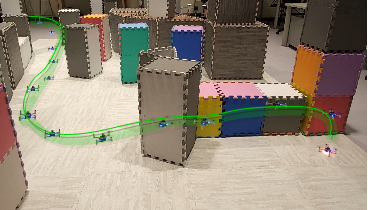}
         \label{fig:renv1}}
  \subfigure[CN-$Q_t$ Path]  	
    {\includegraphics[height=1.2in, width=1.65in]
    {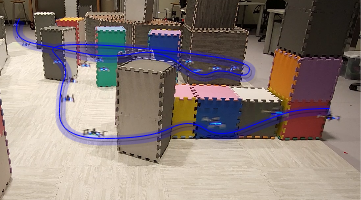}
         \label{fig:renv2}}
  \subfigure[CI-$Q_t$ Path (Failed)]
  	{\includegraphics[height=1.2in, width=1.65in]
    {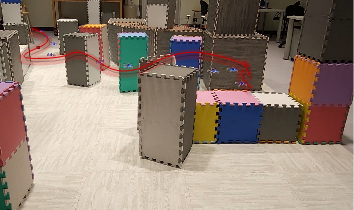}
         \label{fig:renv3}}
    \caption{Illustrates the experimental setup and snapshots of drones in action. (a) Depicts the initial environment. Drone starts from green circle and goal is represented by light blue circle. (b) Shows the snapshots drone following the path by our proposed method. (c) Demonstrates the path for CN-$Q_t$. (d) shows the path for CI-$Q_t$. Note that the testing environment is different than the initial environment. 
    \vspace{-15pt}
    }    
    \label{fig:real-exp}
\end{figure*}
\subsection{The Crucial Role of Initial Context}
In the domain of map prediction, the significance of the initial context cannot be overstated, as it serves as a critical determinant of reliability. To underscore this, we conducted a comprehensive experiment, wherein we progressively revealed the map's details from $0\%$ to $80\%$. Our observations of the resulting impact on both the distance to the goal and navigation efficiency revealed compelling insights.
As depicted in Fig.~\ref{fig:dist2goal2}, the distance to the goal exhibited a consistent improvement with each increment in the initial context provided to the proposed method. This, in turn, translated into enhanced navigation efficiency, as illustrated in Fig.~\ref{fig:nav_eff2}. To fortify our findings, we conducted this experiment across four distinct maps as shown in Fig.~\ref{fig:B3}, consistently yielding shorter path lengths. Note that  the path length for a failed trial refers to the total distance traveled before the termination of that particular trial. The path lengths further reduced given more initial context (visualised by the vertical black line on each bar in Fig.~\ref{fig:B3}). 
The path followed by the all the methods are visualised on the maps for better understanding of the performance comparison. In absence of complex obstacles the performance of CN-$Q_t$ and CG-$Q_t$ are the same as shown in the Fig.~\ref{fig:env3}. The CI-$Q_t$ fails to achieve the goal in all scenarios as they completely ignored the context. In general, the proposed methods avoids complex obstacles as shown in Fig.~\ref{fig:env0} to Fig.~\ref{fig:env3}, however, its performance is influenced by the inital context. This type of initial context is often encapsulated in the form of well-known landmarks - a resource readily available in real-world scenarios. Our work, therefore, offers a systemic approach to incorporating landmark information into planning within the bounds of rationality. This practical approach to navigation showcases the relevance of the proposed work to real-world applications and sets the stage for future advancements.
The readers are encouraged to watch the experimental videos at \url{https://www.youtube.com/watch?v=Ial-EDNPSe0}.

\section{Physical Experiments}
In this section, we show the results of our physical experiments, designed to validate the adaptability and robustness of our proposed method, leveraging the agile Crazyflie 2.1 nano-drones within the motion capture system~\cite{crazyswarm}. We incorporated an observation model with a restricted sensing range to replicate navigation within an unfamiliar map scenario. Our experimental environment recreates real-world challenges by introducing additional obstacles to the initial map that was similar to the maps in the training set. We deliberately added obstacles along the path to the goal. This deliberate perturbation of the environment serves as a litmus test for the adaptability and resilience of our planner, particularly when confronted with maps that slightly deviate from the training distribution.
As shown in Fig.~\ref{fig:real-exp}, all methods adeptly navigated around the obstacles, a testament to the adaptiveness inherent in the bounded rationality framework. However, our approach takes the optimal path to the goal due to a more informative default policy designed to consider predicted map. This transition from simulation to real-world demonstrate the consistency and robustness of our method.

\section{Conclusion}
This paper presents a `context-generative default policy' to address autonomous navigation in unknown environments by leveraging map prediction to anticipate and proactively avoid unseen obstacles.
The adaptive nature of the bounded-rationality framework allows the robot to effectively handle unreliable predictions by selectively sampling trajectories in the vicinity of the default policy. 
Extensive evaluations have demonstrated the superiority of the proposed work. 
This work presents a systematic approach to integrate context into the planning process, paving the way for future research in which visual inputs in the form of context can be incorporated into the map prediction module. 



\bibliography{ref} 
\bibliographystyle{ieeetr}

\end{document}